\documentclass[letterpaper, 10 pt, conference]{ieeeconf}  %

\IEEEoverridecommandlockouts                              
\usepackage{times}
\usepackage{multicol}
\usepackage{graphicx}
\usepackage{subfigure}
\usepackage{epsfig}
\usepackage{amsmath,amssymb,bm}
\usepackage{algorithmic}
\usepackage{url}
\usepackage{color}
\usepackage[ruled]{algorithm2e}
\usepackage{ntheorem}
\usepackage{float}
\usepackage{physics}
\usepackage{cite}
\usepackage{todonotes}
\usepackage{mathtools}

\usepackage[normalem]{ulem}
\usepackage{svg}
\usepackage{soul}

\usepackage{amsmath}

\DeclareMathOperator*{\argmin}{arg\,min}

\title{\LARGE \bf
Learning Decentralized Flocking Controllers with \\ Spatio-Temporal Graph Neural Network
}

\author{Siji Chen$^{1}$, Yanshen Sun$^{1}$, Peihan Li$^{2}$, Lifeng Zhou$^{2}$, Chang-Tien Lu$^{1}$ % <-this % stops a space
\thanks{$^{2}$Siji Chen, Yanshen Sun, and Chang-Tien Lu are with the Department of Computer Science, Virginia Tech, Falls Church, VA 22043, USA. Email: \texttt{\small \{sijic, yansh93, ctlu\}@vt.edu}.}
\thanks{$^{2}$Peihan Li and Lifeng Zhou are with the Department of Electrical and Computer Engineering, Drexel University, Philadelphia, PA 19104, USA. Email: \texttt{\small \{pl525,lz457\}@drexel.edu}.}%
}
\begin{document}

\maketitle
\thispagestyle{empty}
\pagestyle{empty}

\begin{abstract}
Recently a line of researches has delved the use of graph neural networks (GNNs) for decentralized control in swarm robotics. However, it has been observed that relying solely on the states of immediate neighbors is insufficient to imitate a centralized control policy. To address this limitation, prior studies proposed incorporating $L$-hop delayed states into the computation. While this approach shows promise, it can lead to a lack of consensus among distant flock members and the formation of small clusters, consequently resulting in the failure of cohesive flocking behaviors. 
Instead, our approach leverages spatiotemporal GNN, named STGNN that encompasses both spatial and temporal expansions. The spatial expansion collects delayed states from distant neighbors, while the temporal expansion incorporates previous states from immediate neighbors. The broader and more comprehensive information gathered from both expansions results in more effective and accurate predictions.
We develop an expert algorithm for controlling a swarm of robots and employ imitation learning to train our decentralized STGNN model based on the expert algorithm. We simulate the proposed STGNN approach in various settings, demonstrating its decentralized capacity to emulate the global expert algorithm.
Further, we implemented our approach to achieve cohesive flocking, leader following and obstacle avoidance by a group of Crazyflie drones. The performance of STGNN underscores its potential as an effective and reliable approach for achieving cohesive flocking, leader following and obstacle avoidance tasks.
\end{abstract}

\section{Introduction}

Flocking is a collective behavior observed in groups of animals or autonomous agents, such as birds, fish, or artificial robots, where individuals move together in a coordinated manner. In a flock, each robot follows simple rules based on local information to achieve a common group objective~\cite{reynolds1987flocks}. 
Multi-robot systems based on flocking models exhibit self-organization and goal-directed behaviors, making them suitable for various applications, including automated parallel delivery, sensor network design, and search and rescue operations~\cite{majid2022swarm}.
Flocking is typically modeled as a consensus or alignment problem, aiming to ensure that all robots in the group eventually agree on their states~\cite{olfati2006flocking}. Classical methods such as those proposed by Tanner~\cite{tanner2003stable} and Olfati-Saber~\cite{olfati2006flocking}, define rules and constraints governing the position, speed, and acceleration of the robots. 
However, these methods heavily rely on parameter tuning and are limited to predefined scenarios. In contrast, learning-based methods spontaneously explore complex patterns and adapt their parameters through training, providing more flexibility and adaptability compared to classical approaches.

\begin{figure}[h]
    \centering
    \includegraphics[width=\linewidth]{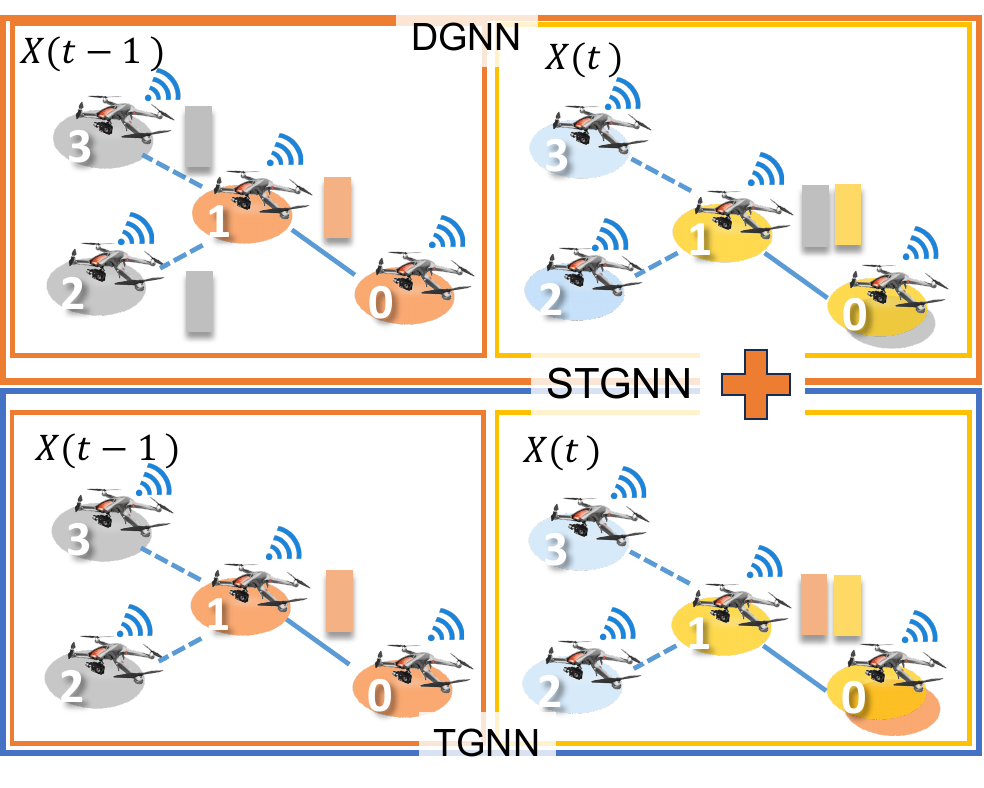}
    \caption{
    The comparison of the information gathered by DGNN (top) and TGNN (bottom) in the case of four aerial robots.
    DGNN propagates the states of robots 2 and 3 at $t-1$ to robot 0 through robot 1, along with robot 1's current state at $t$.
    % However since the communication link is broke at $t_{-1}$, robot 0 can't receive robot 1's $t_{-1}$ state. 
    % Since robot 1 is adjacent to node 0, the history state is as import as delayed state from faraway neighbors. 
    TGNN propagates the states at the previous step $(t-1)$ along with the current state of the robot 1 at $t$ to robot 0.
    STGNN combines information from both DGNN and TGNN, thus processing superior predictive power than either method alone.
    }
    \label{fig:intro}
\end{figure}
There are primarily two research directions in learning-based methods. One approach focuses on imitation learning, as demonstrated by Tolstaya et al.~\cite{Tolstaya19-Flocking}, Kortvelesy et al.~\cite{kortvelesy2021modgnn}, Zhou et al.~\cite{zhou2022graph}, and Lee et al.~\cite{lee2023graph}.  
% Imitation learning has proven to be effective in addressing various problems, including flocking problems~\cite{swamy2022sequence}.
The other approach involves multi-robot deep reinforcement learning (MADRL), as explored in the studies of Yan et al.~\cite{yan2023collision} and Xiao et al.~\cite{xiao2023graph}. MADRL is particularly useful when labels are unavailable, but it presents its own set of challenges, including the demand for an extensive volume of training data and limitations in generalizing to new and unencountered scenarios~\cite{orr2023multi}.
In this work, we choose to utilize imitation learning due to the availability of an expert policy that has proven to be effective for our task~\cite{tanner2003stable, Tolstaya19-Flocking, swamy2022sequence}. 
Recent research in this direction adopts a graph-based approach to represent robots and their interactions and leverages Graph Neural Networks (GNNs)~\cite{hamilton2017inductive, kipf2016semi} for modeling and analyzing flock dynamics. This approach shows promise in addressing flocking tasks by harnessing the power of graph-based representations and neural networks. Specifically, studies such as Tolstaya et al.~\cite{Tolstaya19-Flocking}, Kortvelesy et al.~\cite{kortvelesy2021modgnn}, Zhou et al.~\cite{zhou2022graph}, and Lee et al.~\cite{lee2023graph} utilize a technique called ``delayed state" to incorporate the information from the $l$-step-before states of a robot's $l$-hop neighbors, where $l=1,2,\cdots, L$~\cite{Tolstaya19-Flocking}. Henceforth, we identify this type of model as ``delayed graph neural network'' (DGNN). DGNN enables the learning of spatially extended representations in the local network. However, it overlooks the influence of a robot's historical states and the historical states of its neighbors, thereby neglecting the temporal sequence of flock movement. Diffidently, the spatial and temporal expansion of STGNN enables it to gather information from both spatial and temporal dimensions. We illustrate the distinct information collected by DGNN and TGNN, as well as their combined information gathered by STGNN in Figure~\ref{fig:intro}.
With this insight, we make the following primary contributions in this paper.
\begin{itemize}
    \item \textbf{Design a STGNN-based imitation learning framework for decentralized flocking control.} 
    STGNN enables effective information fusion by integrating delayed states from distant neighbors and previous states from immediate neighbors. To the best of our knowledge, we are the first to design a STGNN-based learning framework for multi-robot flocking with leader following and obstacle avoidance.

    \item \textbf{Develop a centralized expert algorithm for flocking, with leader following and obstacle avoidance.} Finding an expert algorithm is the key to success in imitation learning. We develop an expert algorithm that provides effective control over a large swarm of robots based on Tanner's~\cite{tanner2003stable} and Olfati-Saber's work~\cite{olfati2006flocking}. To the best of our knowledge, we are the first to offer a complete global expert algorithm, capable of handling the flocking with both leader following and obstacle avoidance. 
\item \textbf{Conduct an extensive evaluations of STGNN.} We comprehensively evaluate STGNN by comparing it to 
DGNN and TGNN and testing STGNN with varying history horizons. The results demonstrate STGNN's effectiveness and superior performance in completing complex flocking tasks.
Additionally, we implement STGNN into a group of Crazyflie drones to achieve flocking with obstacle avoidance.
\end{itemize} 
\section{Problem Formulation}
\label{sec:problem_formulation}
% Consider there are $N$ robots on a planar. We assume that all robots are identical with the same ability: max acceleration $u_{max}$, max velocity $v_{max}$, and sensing range $R_c$.  Each robot $i$ is identified by its position $\mathbf{p}_i$ and its velocity $\mathbf{v}_i=\mathbf{\dot p}_i$. The objective of the model is to propose the next acceleration command, $\mathbf{u}_i=\mathbf{\dot v}_i$ of each robot based on the state $(\mathbf{p},\mathbf{v})$. Each robot $i$ is a vertex $V_i$ in the network $\mathcal{G}$. An edge $(i,j)$ exist when the distance $d_{i,j} < R_c$. An adjacency matrix $A \in R^{N\times N}$ where $(j,i) \in \mathcal{E}$.
Consider a collection of $N$ robots that are identical and possess the same capabilities such as maximum acceleration $U_{\text{max}}$, maximum velocity $V_{\text{max}}$, and communication range $R_c$. Each robot $i$ can be uniquely identified by its position $\mathbf{p}_i$ and its velocity $\mathbf{v}_i$.
% , where the velocity can be represented as the time derivative of the position, i.e., $\mathbf{v}_i = \frac{d\mathbf{p}_i}{dt}$.
The task for our learning model is to compute the control input $\mathbf{u}_i$, i.e., acceleration for each robot, based on the current state by itself and its neighbors, represented by the position and velocity $(\mathbf{p},\mathbf{v})$. 
% \YS{In this case, the robot swarm can operate as the objectives required. }

In this paper, we aim to solve three problems as a whole: robot flocking, leader following, and flocking with obstacle avoidance. Each robot is required to avoid collisions with other robots, follow and maintain proximity to a virtual leader, and navigate around obstacles. 
% In our model, we consider each robot $i$ as a node denoted by $V_i$ in the network $\mathcal{G}$. An edge $(i,j)$ exists between two robots if the distance $r_{i,j}$ between them is less than $R_c$. This connectivity can be captured using an adjacency matrix $A \in \mathbb{R}^{N\times N}$, where an element $A_{i,j}$ is non-zero if and only if the edge $(i,j)$ is present in the network $\mathcal{G}$.
Specifically, flocking is to maintain a consistent distance from other robots while synchronizing their movements. The leader following asks the robot swarm to track one or more leader(s) during flocking.
% A virtual leader that governs the desired motion of the flock is characterized by its position $\mathbf{p}^\gamma$ and velocity $\mathbf{v}^\gamma$.
We opt for a single virtual leader for the entire swarm, as the primary objective of flocking is to achieve consensus, and introducing multiple leaders could potentially violate this objective. It is important to note that the virtual leader differs from other robots in that it does not need to avoid collisions. Instead, it represents a predefined trajectory known to all robots. Finally, the obstacle avoidance requests the robot swarm to achieve the two tasks above while avoiding crashing into obstacles. 
\section{METHODOLOGY}
\label{sec:method}
In this research, we employ imitation learning to train STGNN. Specifically, we train an expert model whose outputs serve as labels for training our STGNN model. The expert assumes that the strategy provider possesses real-time information of all robots (i.e., centralized communication), which is impracticable in reality. In contrast, STGNN considered real-world scenarios, where each robot makes decisions by itself with local communication and delayed information (i.e., decentralized communication). Due to the limited information access in decentralized scenarios, STGNN may not generate strategies as effectively as the expert. Nevertheless, after being trained with historical and neighboring data and the output of the expert model, STGNN demonstrates the capability to generate strategies that approach expert-level optimization, despite the limited input data.
% In this section, we discuss the construction of an expert algorithm, which is a centralized flocking approach inspired by Tanner~\cite{tanner2003stable} and Saber~\cite{olfati2006flocking}. Then, we introduce the proposed STGNN-based learning framework, its architecture, and its flexibility to extend to various configurations.
\subsection{Problem Statement}
Denote the robot network at timestamp $t$ as $\mathcal{G}^{(t)}(\mathbf{V}^{(t)}, \mathbf{E}^{(t)}, \mathbf{X}^{(t)}$), where $\mathbf{V}^{(t)}$ is the set of robots, $\mathbf{E}^{(t)}$ is the set of direct communications between nearby robots, and $\mathbf{X}^{(t)}$ is the set of states for the robots. For an arbitrary node $V_i$, its neighborhood $\mathcal{N}_i^{(t)}$ at timestamp $t$ is composed of other robots $V_j$ within its communication range $R_c$, i.e., $\{V_j \in \mathcal{N}_i^{(t)}| \forall V_j, r_{i,j}^{(t)}<R_c\}$. $r_{i,j}^{(t)}$ is the distance between $V_i$ and $V_j$ at $t$. The problem can then be summarized as below:
\begin{equation}
\label{eq:expert_problem}
f_{\text{expert}}(X_i^{(t)}, \mathbf{V}^{(t)}, \mathbf{E}^{(t)}) \rightarrow \mathbf{u}^{(t+1)}, 
\end{equation}
\begin{equation}
\label{eq:stgnn_problem}
\begin{split}
f_{\text{STGNN}}(&X_i^{(t-L)}...X_i^{(t)}, \mathcal{N}_i^{(t-L)}...\mathcal{N}_i^{(t)}, \\
&\mathbf{E}_{i, j; V_j \in \mathcal{N}_i^{(t-L)}...\mathcal{N}_i^{(t)}}, \theta) \rightarrow \mathbf{u}'^{(t+1)},
\end{split}
\end{equation}
\begin{equation}
\label{eq:optimize_problem}
\theta^*= \argmin_{\theta}(\sum_{V_i \in \mathbf{V}} ( \mathbf{u}^{(t+1)}- \mathbf{u}'^{(t+1)})^2),
\end{equation}
where $L$ is the number of historical states used in STGNN, $\theta$ is the set of trainable parameters of STGNN, and $\theta^*$ is the optimized parameter set for STGNN.
\subsection{Expert algorithm} 
\label{sec:expert_policy}
Previous researchers recognized the challenge of communication delays in the early stages of robot flocking studies, leading them to focus on decentralized scenarios. Consequently, only a limited number of prior algorithms have been designed to address centralized scenarios in the context of robot flocking problems. Therefore, to formulate an expert model for targeting our tasks and considering centralized scenarios, we propose a novel centralized model to generate labels for STGNN training.

As introduced in Section~\ref{sec:problem_formulation}, an expert algorithm should cover three crucial aspects---flocking, the leader following, and obstacle avoidance. In this case, the algorithm should be constrained by both the distances between robots and the distances between robots and obstacles. Similar to~\cite {tanner2003stable}, the regularized update of speed is computed jointly from the collision avoidance potential and the velocity agreement. The collision avoidance potential ensures that the distances between robots exceed a predefined threshold, while the velocity agreement ensures that robots maintain consistent behavior in relation to the other robots. 

Specifically, consider there exists a control $u_i$ of robot $i$ as the input of the expert algorithm. The algorithm then generates an update $u_i$ following specific rules.
% the control input $u_i$ generated by the expert algorithm for each robot $i$
$u_i$ can be defined by the combination of these components as in Equation~\ref{eq:expert_control}. In Equation~\ref{eq:expert_control}, $c_\alpha$, $c_\beta$, and $c_\gamma$ are positive weighting parameters. The $\alpha$- term specifies collision avoidance and velocity alignment among the robots, while the $\beta$- term specifies collision avoidance and velocity alignment between robots to obstacles. The $\gamma$ term is peer-to-peer guidance from the virtual leader to each robot $i$.
% The expert control policy can be written as the sum of all three individual control \ref{eq:exper_control}.
\begin{equation}
\label{eq:expert_control}
 \mathbf{u}_i=c_{\alpha} \mathbf{u}_i^\alpha+c_{\beta} \mathbf{u}_{i}^\beta+c_{\gamma} \mathbf{u}_{i}^\gamma.
\end{equation}

% Tanner et al.~\cite{tanner2003stable} proposed to generate the acceleration command $u^\alpha$ by combining the collision avoidance potential (as defined in Equation \ref{eq:flocking_u_distance}) with the velocity agreement within the flock (as the first term defined in Equation \ref{eq:flocking_u_v}).
The collision avoidance potential $U$ increases significantly
% plays a crucial role in ensuring that the robots maintain a safe distance from each other. 
as the distance $r_{i,j}$ between two robots decreases.
% becomes smaller, the term $U_{i,j}$, representing the collision avoidance potential, increases significantly. 
This increment is governed by a reciprocal function that dominates when $r_{i,j}$ approaches zero (Eq.~\ref{eq:flocking_u_distance}). Consequently, $U_{i,j}$ captures the fact that beyond a certain distance threshold (e.g., communication range $R_c$), no direct interaction exists between robots in terms of collision avoidance. The velocity agreement is the velocity difference between robot $i$ and all the other robots (the first term in Eq.~\ref{eq:flocking_u_v}).
% In addition, by incorporating velocity alignment with collision avoidance, we enhance our centralized expert control mechanism.
The resulting control input $u_i^\alpha$ (as defined in Eq.~\ref{eq:flocking_u_v}) is a centralized solution that takes into account the velocity mismatch among all robots and the local collision potential. Note that the collision avoidance implemented by Equation \ref{eq:flocking_u_distance} does not guarantee a minimal distance between robots. Our results show that the minimal distance decreases when the number of robots increases.
% and possesses comprehensive knowledge of all robots within the system.
\begin{equation}
\label{eq:flocking_u_distance}
    U_{i,j}= \frac{1}{r_{i,j}^2} + \texttt{log}||r_{i,j}||^2, ||r_{i,j}||\leq R_c. \\
    % U_{i,j}= \begin{cases}
    %     &\frac{1}{r_{ij}^2} + log||r_{ij}||^2, ||r_{ij}||\leq\rho \\
    %     &0, ||r_{ij}||>\rho 
    % \end{cases}
\end{equation}
\begin{equation}
    \label{eq:flocking_u_v}
     \mathbf{u}_i^\alpha=-\sum_{j=1}^N( \mathbf{v}_i- \mathbf{v}_j)-\sum_{j=1}^N(\nabla_{r_{i,j}}U_{i,j}).
\end{equation}
% For leader following and obstacle avoidance tasks, we extends the solution from Olfati-Saber \cite{olfati2006flocking}. Olfati-Saber's solution defines three types of agents: $\alpha$-agent is member of the swarm, $\beta$-agent is the kinematic agent induced by an $\alpha$-agent in close proximity of an obstacle, and $\gamma$-agent is a navigational feedback which will lead to the direction of the trajectory.
To incorporate obstacle avoidance in the model, we follow Olfati-Saber~\cite{olfati2006flocking}'s work by introducing an imaginary robot, i.e., $\beta$-robot, which is defined by the projection of a robot $i$ on the $k$-th obstacle $O^k$ within its communication distance $R_c$, to assist with the obstacle avoidance task. In practice, this can be achieved by a robot, equipped with sensors,  measuring the relative position and velocity between the closest point on an obstacle and itself~\cite{olfati2006flocking}. The control input $\mathbf{u}_i^\beta$ follows the flocking control (Eq.~\ref{eq:flocking_u_v}) while focusing on the potential between robot $i$ and its projection on obstacles.
Let $\mathbf{p}_k^o$ denote the position of the $k$-th obstacle. Then the position and velocity of the $\beta$-robot, created by projecting the $i$-th robot on the $k$-th obstacle, can be calculated by Equation \ref{eq:obs_proj} and Equation \ref{eq:obs_proj_v}, respectively.
\begin{equation}
    \label{eq:obs_proj}
     \mathbf{p}_{i,k}=\mu*\mathbf{p}_i+(1-\mu)\mathbf{p}_k^o, \; \mu=\frac{r_k}{||\mathbf{p}_i-\mathbf{p}_k^o||}.
\end{equation}
\begin{equation}
    \label{eq:obs_proj_v}
     \mathbf{v}_{i,k}=\mu P\mathbf{v}_i, \; P=I-\mathbf{a}_k \mathbf{a}_k^T, \; a_k=\frac{(\mathbf{p}_i-\mathbf{p}_k^o)}{||\mathbf{p}_i-\mathbf{p}_k^o||}.
\end{equation}
% \begin{figure}
%     \centering
%     \includegraphics[width=0.4\linewidth]{figures/proj_fig1.pdf}
%     % {figures/proj_plot_1_obs_1x.png}
%     \caption{$\beta$-robot representation on a spherical obstacle. 
%     }
%     \label{fig:obs_projection}
% \end{figure}
Define the state of the virtual leader by its position $p^r$ and velocity $v^r$. The leader following control $u^r$ can be defined in Equation \ref{eq:gamma_control}, where $c_1$, $c_2$ are positive weighting parameters.
\begin{equation}
    \label{eq:gamma_control}
    \mathbf{u}_i^\gamma=-c_1(\mathbf{p}_i-\mathbf{p}^r)-c_2(\mathbf{v}_i-\mathbf{v}^r), c_1>0, c_2=\sqrt{c_1}.
\end{equation}

\subsection{STGNN-based imitation learning }
In this section, we present an overview of the STGNN-based learning model.  
% functions as a robot flocking action prediction model. 
We begin by describing the strategy used to construct the model input. Then we discuss the architecture of the STGNN model, which involves two levels of state expansion.

\begin{figure}[h]
  \centering
  \includegraphics[width=\linewidth]{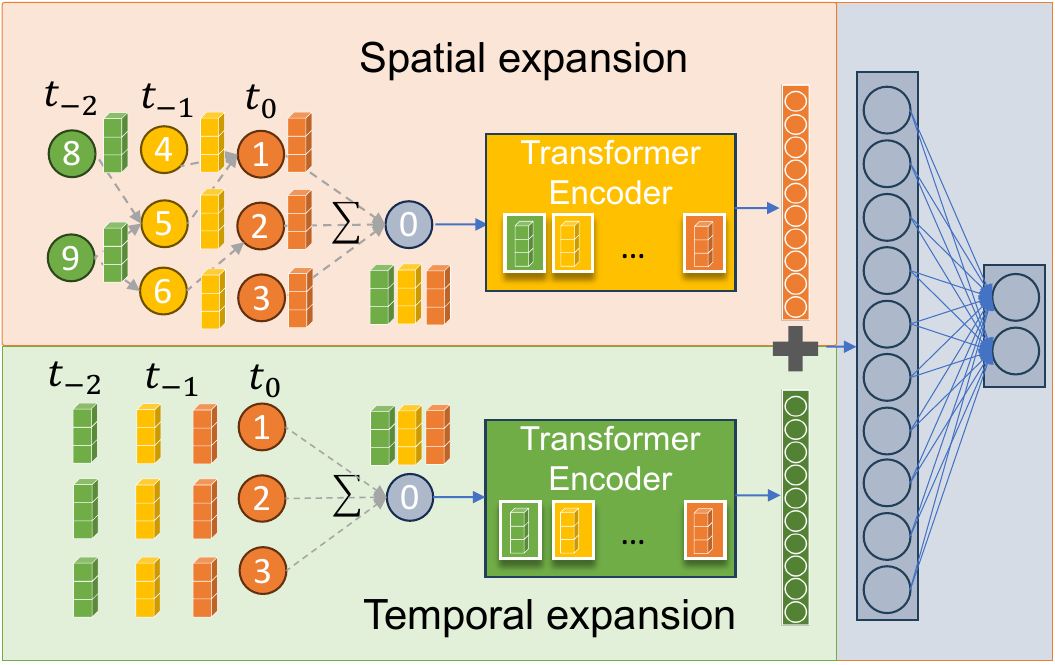}
 % \centering
 %  \includegraphics[width=0.9\linewidth]{figures/temporal_exp.pdf}
 
   \caption{STGNN spatial and temporal expansion module with $L=3$. Spatial expansion and temporal expansion operate in parallel. 
   % In spatial expansion, $L$-hop delayed state is propagated when a pair of robots are within communication range. The sequence of delayed states are passed to the transformer encoder to complete the spatial expansion. Meanwhile the history states of immediate neighbors are passed to transformer encoder to complete the temporal expansion. 
   Both spatial and temporal expanded states are concatenated and passed to the next layer of the neural network to predict the control input.}
  \label{fig:spatial_temporal_expansion}
\end{figure}

\paragraph{Local observation}
\label{sec:method_local_observation}
We develop a decentralized solution that only requires local observations of each robot. Inspired by Tolstaya \cite{Tolstaya19-Flocking}, we define the local state of $i$-th robot in Equation~\ref{eq:x_i_input_all3}, which consists of the aggregation from neighboring robots ($\alpha$ term), local observation of obstacles ($\beta$ term), and local observation of the virtual leader ($\gamma$ term). The local observation of obstacles is defined as $k \in \mathcal{N}$ with $r_{i,k}<R_c$. $r_{i,k}$ is the distance between robot $i$ and $\beta$-robot, the projection of robot $i$ on the $k$-th obstacle.
\begin{equation}
\label{eq:x_i_input_all3}
    X_i^{\alpha,\beta,\gamma}=[ [\sum_{j\in \mathcal{N}}{X_{i,j}^{\alpha}}]; [\sum_{k \in \mathcal{N}}{X_{i,k}^{\beta}}]; [X_{i}^{\gamma}]].
\end{equation}
Instead of directly using the position $\mathbf{p}_i$ and velocity $\mathbf{v}_i$ of robot $i$ as input for our model, we adopt the relative state to be consistent with the expert algorithm. We define the relative state for robot $i$ to robot $j$ as $X_{i,j}$ (Eq.~\ref{eq:x_i_input}). Then we aggregate the local peer-to-peer relative states into a local state $X_i$, which includes $X_i^\alpha$ (Eq.~\ref{eq:x_i_input_agg_alpha}) and $X_i^\beta$ (Eq.~\ref{eq:x_i_input_agg_beta}).

\begin{equation}
    \label{eq:x_i_input}
    X_{i,j}=[\mathbf{v}_i-\mathbf{v}_j,\frac{\mathbf{p}_i-\mathbf{p}_j}{r_{i,j}},\frac{\mathbf{p}_i-\mathbf{p}_j}{r_{i,j}^2}].
\end{equation}
\begin{equation}
    \label{eq:x_i_input_agg_alpha}
    X_i^\alpha= \frac{1}{\mathcal{N}} \sum_{j \in \mathcal{N}}X_{i,j}^\alpha, \; j\in \mathcal{N} \; \textrm{if} \; r_{i,j}<R_c.
\end{equation}
\begin{equation}
    \label{eq:x_i_input_agg_beta}
    X_i^\beta= \sum_{k \in \mathcal{N}}X_{i,j}^\beta, \; k\in \mathcal{N} \; \textrm{if} \; r_{i,o}<R_c.
\end{equation}
Notably, our model only requires local state information defined by Equation \ref{eq:x_i_input}, which makes our approach decentralized. 
\paragraph{STGNN}
In order to mimic the control input $\mathbf{u}$ generated by the expert algorithm that uses the global information, we implement two levels of expansions based on local observation as illustrated in Figure~\ref{fig:spatial_temporal_expansion}. The first level is spatial expansion, aiming at extracting information from additional robots, while the second level is temporal expansion, which centers on local state evolution. For each expansion, we extract information on $L$ timestamps. For the timestamp $t$, consider an arbitrary node $V_i$ with state $X_i^{(t)}$ and $1$-hop neighborhood $\mathcal{N}_i^{(t)}$. A node $V_j$ is $V_i$'s $l$-hop neighbor if it is a $1$-hop neighbor of at least one $V_i$'s $l-1$-hop neighbor. Note that even though we use ``$l$-hop neighbor" to account for state delays resulting from spatial distances, the state of one hop further is the same as one timestamp earlier in our scenario. In other words, the state of a $l$-hop neighbor $j$, $X_j^{(l)}$, is equivalent to $X_j^{(t-l+1)}$.

In spatial expansion, we merge each node's current state $X_i^{(t)}$ to its $l$-hop delayed neighbors, where $l=1, 2, ..., L$. Specifically, we first aggregate the delayed states for each hop, as indicated in Equation~\ref{eq:k_delayed_state}. Then, $X_i^{(t)}$ is updated with its neighborhood information as in Equation~\ref{eq:t_delayed_state}.
% $X_k$ represents the delayed state at a specific time step in the past (e.g., $t=-k$). We utilize the adjacency matrix $A_k$, which captures the connectivity at time step $t=-k$ to merge in the delayed state (as defined in Equation \ref{eq:k_delayed_state}).
\begin{equation}
    \label{eq:k_delayed_state}
    {X'}_{i\_s}^{(l)}=\sum_{j \in \mathcal{N}_i^{(l)}} X_j^{(l)}, l=1, 2, ..., L.
\end{equation}
\begin{equation}
    \label{eq:t_delayed_state}
    {{H}_{i\_s}^{(t)}} = [{X}_i^{(t)} ||_{l=1, ..., L} {X'}_i^{(l)}].
\end{equation}
The same method is applied to other timestamps $l=t-L, ..., t-1$, but with $t-l+1$-hop neighbors. Then, $H_{i\_s}=\{{{H}_{i\_s}^{(l)}}, l=t-L, ..., t\}$ are fed into a transformer~\cite{vaswani2017attention} for temporal-wise information fusion. The transformer takes three copies of $H_{i\_s}$ and considers them as query, key, and value separately. The query and key are used to compute relations between different timestamps and the value is used to generate temporal-wise fused embeddings:
\begin{equation}
\label{eq:transformer}
{H'}_{i\_s} = \sigma(\frac{WqH_{i\_s}(WkH_{i\_s})^T}{\sqrt{d}})WvH_{i\_s}
\end{equation}
where $\sigma$ is a softmax function and $d$ is the number of features of ${H_i}_s$.

% Finally, the output ${S}_i^{(t)}$ of spatial expansion is then considered as the input of the temporal expansion.

% \SJ{\begin{equation}
%     \label{eq:k_delayed_state_temp}
%     X^l=[\prod_{i=0}^l A^i X^l ]
% \end{equation}
% % \begin{equation}
% %     \label{eq:t_delayed_state}
% %     X_K^t= A_0 X_k.
% % \end{equation}
% \begin{equation}
%     \label{eq:x_input_historyK}
%     H^{(t)} =[X^1;\dots;X^t].
% \end{equation}
% }

% The temporal expansion uses the delayed state $X_i^{0,...,K-1}$, passing it using the latest connectivity $A_K$, 
% Next we pass all the delayed states $X_k$ through the first Long Short-Term Memory (LSTM) layer. The LSTM network allows us to capture the delayed states of $K$-hop neighbors, expanding the spatial representation of the system. The spatially expanded embeddings are then stored sequentially and passed through the second LSTM layer for temporal expansion. The temporal expansion plays a crucial role in predicting the movement of the swarm, overcoming the limitations imposed by relying solely on local information. This process is illustrated in Figure \ref{fig:spatial_temporal_expansion}. 
The temporal expansion considers the temporal evolution pattern for local spatial state fusion (i.e., considers only $1$-hop neighborhood). Specifically, the spatial data fusion method is shown in Equations~\ref{eq:k_delayed_state_t} and~\ref{eq:t_delayed_state_t}, and the temporal feature fusion is achieved by a transformer as described in Equation~\ref{eq:transformer}. Consider the output of the temporal expansion module as ${H'}_{i\_t}$, the concatenation of ${H'}_{i\_s}$ and ${H'}_{i\_t}$ is fed to a feed-forward network to generate the final flocking controls for the robots.

\begin{equation}
    \label{eq:k_delayed_state_t}
    {X'}_{i\_t}=\sum_{j \in \mathcal{N}_i^{(1)}} X_j^{(l=1)}.
\end{equation}
\begin{equation}
    \label{eq:t_delayed_state_t}
    {{H}_{i\_t}} = [{X}_i^{(t)} ||{X'}_i^{(l=1)}].
\end{equation}

% By employing temporal expansion, the model can incorporate past information and enhance the understanding of the system's dynamics. Both $X_K^s$ and $X_K^t$ are concatenated as input $X_K$ of timestamp $t=t_0$ with history length equal to $K$. The expansion of spatial and temporal states is passed through one layer transformer to extract the hidden states to make prediction of $\hat{u}$.

% \begin{figure}[h]
%   \centering
%   \includegraphics[width=\linewidth]{figures/temporal_exp.pdf}
%   \caption{STGNN model temporal expansion}
%   \label{fig:temporal_expansion}
% \end{figure}

\section{Simulation and experiment}
\label{sec:experiment_results}
In this section, we discuss the settings of the STGNN-based learning model, evaluation metrics, simulation results, and a real robot experiment. A video of our simulation and experiment is available online.~\footnote{~\url{https://youtu.be/Hb7Ofe3IvaA}}

\subsection{Settings for STGNN-based learning model}
\label{sec:experiment_setup}

\paragraph{Local observation} 

The local observation (Eq.~\ref{eq:x_i_input_all3}) denoted as $X \in \mathbb{R}^{18}$, is passed through a two-layer multi-layer perceptron (MLP) with a hidden size of 128 to extract the local feature embedding $H  \in \mathbb {R}^{128}$ prior to undergoing spatial and temporal expansion. Each $X$ from the historical data follows the same process.

\paragraph{STGNN spatial expansion}
A single SAGEConv layer with a hidden size of 128 is used to aggregate local neighbor information. Then the $l$-hop neighbor information is obtained using Equations \ref{eq:k_delayed_state} and \ref{eq:t_delayed_state}. The result is a sequence of $L$ delayed states. A transformer consisting of two encoder layers, each with a feedforward size of 16 and a head size of 4, is used to extract the spatial expansion results. 
\paragraph{STGNN temporal expansion} 
For each local features from the history $L$, a single SAGEConv layer with hidden size of 128 is used to aggregate neighbor information (Eq. \ref{eq:t_delayed_state}). A transformer consisting of 2 encoder layers, each with a feedforward size of 16 and a head size of 4, is used to extract the temporal expansion result.
\paragraph{Action generation} 
The last output from both spatial and temporal expansions is concatenated to get the fusion embedding $H \in \mathbb{R}^{256}$. Subsequently, this output undergoes a two-layer MLP to generate the predicted control signal $\mathbf{\hat{u}}$. 
The ground truth of the training data set is derived by the expert algorithm (Sec.~\ref{sec:expert_policy}). 
% Our model's objective is to replicate the control signal $u$ generated by the expert algorithm based on local states. 
L2 loss with the Adam optimizer is used to train the model.
% The architecture of our model includes two layers of MLP to aggregate local observations of the states of neighboring robots, the leader, and obstacles and a single SAGEConv layer with a parameter size of 64 to aggregate neighboring states. The aggregated states are then used to compute delayed states using Meanwhile, the local states are used to compute the temporal delayed states \rev{using Equation \ref{eq:}?}. Both types of delayed states are concatenated and sequentially passed to a Transformer layer to extract the fused information. Finally, the output from the transformer is extracted and passed through a two-layer MLP to generate the control input $\hat{u}$.

 We trained the STGNN model with $N=20$, which includes three spherical obstacles and one virtual leader. The size of the swarm is chosen to ensure that spatial expansion could encompass a significant number of robots. The virtual leader goes through the obstacles but the robots must avoid collision with obstacles while following the virtual leader. The virtual leader moves continuously at a constant speed of 1m/s along the x-axis. 
 The communication range is $R_c=1$ m, and the sampling period is $T_s=0.01$ s. The initial robot positions are randomized, following a uniform distribution within the range of $[0, 0.5R_c\sqrt{N}]$. Initial velocity is $0$ m/s.
% uniformly distributed within the interval $[-10, 10]$, with 10 represents $V_{max}$. 
% The choice of $R_c$, $N$, $V_{max}$, and $U_{max}$ are adjustable parameters, allowing us to simulate various real-world settings and evaluate their limitations. Here, 
The maximum allowed acceleration is $U_{\max}=10~\text{m/s}^2$ and the maximum velocity allowed is $V_{\max}=10~\text{m/s}$ on each axis. 
The safety distance is 0.15 m. If $r_{i,j}$ falls below 0.15 m, the experiment is terminated early.
To ensure a valid initial configuration, $r_{i,j}$ must be greater than 0.15 m.
The training episode has 1200 steps to allow all robots to pass obstacles and form flocking behavior on the other side. One trajectory of the expert algorithm which is used in training is shown in Figure~\ref{fig:exp_result_fig_n20}.

We implement our model using PyTorch and the OpenAI gym framework in Python3.9. The server we use has Intel(R) Xeon(R) W-2133 CPU @ 3.60GHz, NVIDIA QUADRO P5000 GPU, and 32 GB RAM.
% \subsection{Compared algorithms}
% \label{sec:compared_algorithms}
We conduct a comprehensive evaluation of our proposed method, STGNN, with $L$ set to $1$, $2$, $3$.
The compared algorithms include DGNN, TGNN, and Olfati-Saber's decentralized flocking algorithm denoted as Saber~\cite{olfati2006flocking}.
% Additionally, we explore a global variant of STGNN (GSTGNN) where the communication range $R_c$ is set as infinity. 
% As a result, this solution did not yield ideal results in our evaluation.
% In addition to Mean Absolute Error (\textbf{$MAE$}), we use three other metrics to evaluate the performance. Minimal Distance of swarm (\textbf{$D_{min}$}) which is the mean of minimal distance between any two robots of the episode. \textbf{Close to expert policy or larger value is better.}  Velocity alignment (\textbf{$V$}) which is the mean of velocity variance of the swarm during the episode. In the leader following scenario, the variance is computed based on leader. \textbf{The lower value is better.}    Distance to Leader (\textbf{$\tau$}) is used in leader following scenario, measures the mean distance from any robot to the leader. \textbf{The lower value is better.} 
\subsection{Metrics}
\label{sec:metrics}
\begin{enumerate}
    \item Completion Rate ($C\%$): the rate of successfully completed episodes. The episode can be terminated early if any robot hits obstacles or $r_{i,j}$ falls below 0.15 m.
    \textit{The higher value is better.}
    \item Mean Absolute Error (MAE): the mean absolute error between the expert control $\mathbf{u}$ and model prediction $\mathbf{\hat{u}}$.
    \textit{The lower value is better.}
    % \item Minimal Distance of swarm($D_{min}$): The mean of minimal distance between any two robots of the episode.
    % A stable flocking should maintain a safety distance between any two robots. Our model learn the behavior indirectly from the expert policy.
    % In \ref{sec:method} section there is no direct control of this parameter. 
    \item Velocity alignment ($V$): the velocity variance of the swarm at the end of the episode. All robots should be velocity aligned at the end of the episode thus \textit{the lower value is better.}    
    \item Distance to the leader ($\tau$, Eq.~\ref{eqn:average_distance_leader}): the mean distance from any robot to the leader is an auxiliary measure. A large value indicates robots deviate from the leader and move in different direction, resulting in the failure of swarm formation. However a small value indicates a higher risk of collision within the swarm.
    \textit{Close to the expert algorithm is better.}
    % The average distance between the target and all the followers was computed by     
    \begin{align}\label{eqn:average_distance_leader}
        \tau = \frac{1}{nT} \sum_{i=1}^{n}\sum_{t=1}^{T}{\tau_i^t}.
    \end{align}
\end{enumerate}
\subsection{Experiment results}
We train one model for each setting described in  Section~\ref{sec:experiment_setup}. The training consists of 200 epochs with an initial learning rate of 1e-3. We implement early stopping and exponential learning rate decay to prevent overfitting.
% We monitor the training loss in real time and decrease the learning rate to 50\% once the training loss increases by 2\%. The training is completed when the learning rate is less than 1e-6. The training episode has 1000 steps and one obstacle with a radius of 1 is located at the 300th step on the leader trajectory. 
\paragraph{Evaluation on swarm size of 20}
In the first set of experiments, the trained models are tested in the same environment as the training environment, as described in Section~\ref{sec:experiment_setup}. The only difference lies in the initial positions of the robots, which vary due to random initialization.
During the testing phase, the swarm's next state is determined by the model's prediction $\mathbf{\hat{u}}$. 
If an episode is terminated early due to collision, the failed episode's metric (MAE, $V$, $\tau$) are not included in the aggregation for reporting. Each model runs over 20 trials and the mean and standard deviation are reported in Table \ref{tab:multi-robot-flocking-with-obs-results-20}. 
For STGNN L1, L2 and L3 models, the results demonstrate consistently improvement as spatial and temporal expansion increases. By increasing L from 1 to 3 in both spatial and temporal expansion, the completion rate increases from 0.85 to 1.0, the MAE decreases from 3.03 to 1.71, and the velocity variance also decreases from 0.14 to 0.01. The distance to the leader does not consistently decrease, but as explained in Section~\ref{sec:metrics}, a smaller value indicates a higher risk of collision, so both values of 1.03 and 1.06 are acceptable.
When considering DGNN L3 and TGNN L3, both models exhibit improved performance compared to STGNN L1, which uses only local neighbor information. Furthermore, STGNN L3, which is the combination of DGNN L3 and TGNN L3, leads to further performance enhancement.
Olfati-Saber's decentralized solution~\cite{olfati2006flocking} achieves a similar success rate as STGNN L1, as both models utilize only local information. However, the Saber algorithm differs from our expert algorithm, and its formations tend to have larger minimal distances, resulting in a larger $\tau$, which is not directly comparable to the expert algorithm.

\begin{figure}
    \centering
    \includegraphics[width=\linewidth]{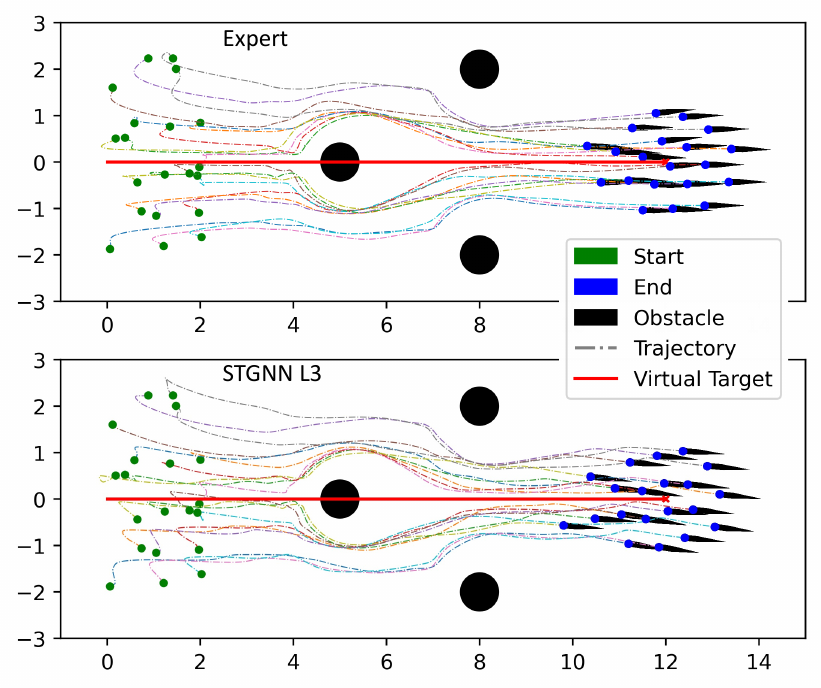}
    \caption{Simulation result of $N=20$. Top: the expert model drives 20 robots. Bottom: STGNN with $L3$ drives 20 robots. The two models are tested in the same environment with the same initialization.}
    \label{fig:exp_result_fig_n20}
\end{figure}

In Figure \ref{fig:exp_result_fig_n20}, a test trajectory of the expert algorithm is plotted on the top and the test trajectory of STGNN L3 is plotted at the bottom. The plots illustrate that STGNN L3 possesses the capability to mimic the expert algorithm's control input $\mathbf{u}$ and generate similar flocking trajectories for the robots. 
% I don't need to explain the Saber's large tau
% However, it's worth noting that the Saber algorithm requires a larger communication range ($R_c$) and more time to achieve formation. Therefore, in this experiment, we set Saber's $R_c$ to 5m and allow 3000 steps to complete the episode. The result $\tau$ of Saber algorithm is scaled back to make it comparable to other models.
% To compare the performance of our ST-GNN approach, we varied the memory settings from 1, 2, 3, to 5. We compared these different memory configurations with baseline models, evaluating their performance using the metrics described in Section \ref{sec:metrics}.
% \paragraph{Multi-robot Flocking} 
%%%%%%%%%%%%%%%%%%%%     multi robot flocking with obs table 1  with 20 agents %%%%%%%%%%%%%%%%%%%%%    
\begin{table}[!h]
\begin{center}
\begin{tabular}{c | c c c c  }
\hline
Model & C\% & MAE & V & $\tau$  \\ 
% \hline
% Expert      & 1.00 & --             &  0.05 $\pm$  0.03 & 5.14 $\pm$ 0.02   \\ 
% Saber      & 0.60 & --             &  0.10 $\pm$  0.13 & 6.37 $\pm$ 0.12   \\ 
% STGNN L1 & 0.80 & 20.47 $\pm$ 4.74 &  0.76 $\pm$  0.94 & 10.01 $\pm$ 5.00   \\ 
% STGNN L2 & 0.90 & 13.62 $\pm$ 4.58 &  0.50 $\pm$  0.50 & 5.84 $\pm$ 0.81   \\ 
% STGNN L3 & \textbf{1.00} &\textbf{ 7.54 $\pm$ 1.40} & \textbf{ 0.13 $\pm$  0.06 }& \textbf{5.22 $\pm$ 0.11}   \\ 
% \hline
\hline
Expert      & 1.00 & --             &  0.00 $\pm$  0.00 & 0.97 $\pm$ 0.01   \\ 
Saber      & 0.85 & --             &  0.02 $\pm$  0.02 & 2.27 $\pm$ 0.14   \\ 
\hline
STGNN L1 & 0.85 & 3.03 $\pm$ 0.74 &  0.14 $\pm$  0.34 & 1.32 $\pm$ 0.31   \\ 
STGNN L2 & 0.90 & 2.25 $\pm$ 0.19 &  0.02 $\pm$  0.03 & 1.03 $\pm$ 0.05   \\ 
STGNN L3 & \textbf{1.00} & \textbf{1.71 $\pm$ 0.17} &  \textbf{0.01 $\pm$  0.01} & 1.06 $\pm$ 0.06   \\ 
\hline
DGNN L3 & 1.00 & 2.83 $\pm$ 0.29 &  0.03 $\pm$  0.11 & 1.08 $\pm$ 0.12   \\ 
TGNN L3 & 1.00 & 2.11 $\pm$ 0.14 &  0.02 $\pm$  0.03 & \textbf{1.02 $\pm$ 0.05}  \\ 
\hline
\end{tabular}
\end{center}
\caption{Test results of $N=20$. L denotes the history length.
% Each test has random initialization and the trajectory is controlled by tested model. 
Overall STGNN L3 achieves the best performance.} 
\label{tab:multi-robot-flocking-with-obs-results-20}
\end{table}
%%%%%%%%%%%%%%%%%%%%%%%%%%%%%%%%%%%%%%%%%%%%%%%%%%%%%%%%%%  
\paragraph{Evaluation on varying swarm sizes} 
In the second set of experiments, we examine our model's transferability across various swarm sizes, specifically for $N =$ 30, 40, and 50. We directly employ previously trained models from larger swarm sizes and conduct 20 trials for each scenario. The mean and standard deviation for each model and swarm size are reported in Table \ref{tab:multi-robot-flocking-with-obs-results-30-40-50}. To facilitate swarm formation after obstacle avoidance, we increased the maximum episode steps from 1200 to 1500, 1800, and 2100 for swarm sizes $N$ = 30, 40, and 50, respectively.

STGNN L3 successfully achieves flocking with leader following and obstacle avoidance through all testing cases and attains the lowest MAE compared to STGNN models with shorter history. Velocity alignment is achieved in both STGNN L2 and STGNN L3. The large $\tau$ in STGNN L1 indicates the failure in swarm formation.
Our proposed expert algorithm consistently performs well in terms of success rate, velocity alignment, and distance to target, which demonstrates its suitability as the ground truth. Furthermore, the performance of STGNN L3 underscores its ability to provide accurate estimations of the global control inputs using only local information.
%%%%%%%%%%%%%%%%%%%%     multi robot flocking with obs  30,40,50 agents %%%%%%%%%%%%%%%%%%%%%     
\begin{table}[!h]
\begin{center}
\begin{tabular}{c | c c c c  }
\hline
& \multicolumn{4}{c}{N 30}      \\
Model & C\% & MAE & V & $\tau$  \\ 
\hline
Expert      & 1.00 & --             &  0.00 $\pm$  0.00 & 1.17 $\pm$ 0.03   \\ 
STGNN L1 & 0.90 & 4.14 $\pm$ 0.32 &  0.06 $\pm$  0.10 & 2.87 $\pm$ 1.99   \\ 
STGNN L2 & 0.90 & 3.19 $\pm$ 0.17 &  \textbf{0.01 $\pm$  0.00} & 1.11 $\pm$ 0.05   \\ 
STGNN L3 & \textbf{1.00} & \textbf{2.69 $\pm$ 0.15} &  0.01 $\pm$  0.01 & \textbf{1.12 $\pm$ 0.05}   \\ 
% STGNN L3 K & 1.00 & 3.59 $\pm$ 0.22 &  0.11 $\pm$  0.20 & 1.22 $\pm$ 0.16   \\ 
% STGNN L3 T & 0.95 & 2.89 $\pm$ 0.21 &  0.01 $\pm$  0.00 & 1.11 $\pm$ 0.03   \\ 
\hline
& \multicolumn{4}{c}{N 40}      \\

Model &  C\% & MAE & V & $\tau$  \\ 
\hline
Expert      & 1.00 & --             &  0.00 $\pm$  0.00 & 1.29 $\pm$ 0.03   \\ 
STGNN L1 & 0.95 & 5.63 $\pm$ 0.21 &  0.14 $\pm$  0.23 & 8.87 $\pm$ 3.36   \\ 
STGNN L2 & 0.90 & 4.43 $\pm$ 0.11 &  \textbf{0.01 $\pm$  0.01} & 1.16 $\pm$ 0.04   \\ 
STGNN L3 & \textbf{1.00} & \textbf{3.88 $\pm$ 0.14} &  \textbf{0.01 $\pm$  0.01} &\textbf{1.16 $\pm$ 0.02}   \\ 
% STGNN L3 K & 0.95 & 4.54 $\pm$ 0.20 &  0.62 $\pm$  0.38 & 1.73 $\pm$ 0.28   \\ 
% STGNN L3 T & 1.00 & 3.99 $\pm$ 0.15 &  0.01 $\pm$  0.00 & 1.19 $\pm$ 0.07   \\ 
\hline
 & \multicolumn{4}{c}{N 50}      \\

Model &  C\% & MAE & V & $\tau$  \\ 
\hline
Expert      & 1.00 & --             &  0.00 $\pm$  0.00 & 1.39 $\pm$ 0.01   \\ 
STGNN L1 & 0.95 & 6.90 $\pm$ 0.20 &  0.89 $\pm$  1.70 & 21.61 $\pm$ 6.34   \\ 
STGNN L2 & 0.95 & 5.44 $\pm$ 0.10 &  0.01 $\pm$  0.01 & \textbf{1.26 $\pm$ 0.09}  \\ 
STGNN L3 & \textbf{1.00} & \textbf{5.00 $\pm$ 0.10} &  \textbf{0.00 $\pm$  0.00} & 1.23 $\pm$ 0.11   \\ 
% STGNN L3 K & 1.00 & 5.43 $\pm$ 0.30 &  0.89 $\pm$  0.77 & 2.16 $\pm$ 0.59   \\ 
% STGNN L3 T & 1.00 & 4.90 $\pm$ 0.13 &  0.00 $\pm$  0.00 & 1.23 $\pm$ 0.09   \\ 
\hline
\end{tabular}
\end{center}
\caption{Test results: $N=30, N=40, N=50$. L denotes the history length.
Overall STGNN L3 achieves the best performance.} 
\label{tab:multi-robot-flocking-with-obs-results-30-40-50}
\end{table}
%%%%%%%%%%%%%%%%%%%%%%%%%%%%%%%%%%%%%%%%%%%%%%%%%%%%%%%%%%  
\subsection{Real robot experiment}
\begin{figure}
    \centering
    \includegraphics[width=\linewidth]{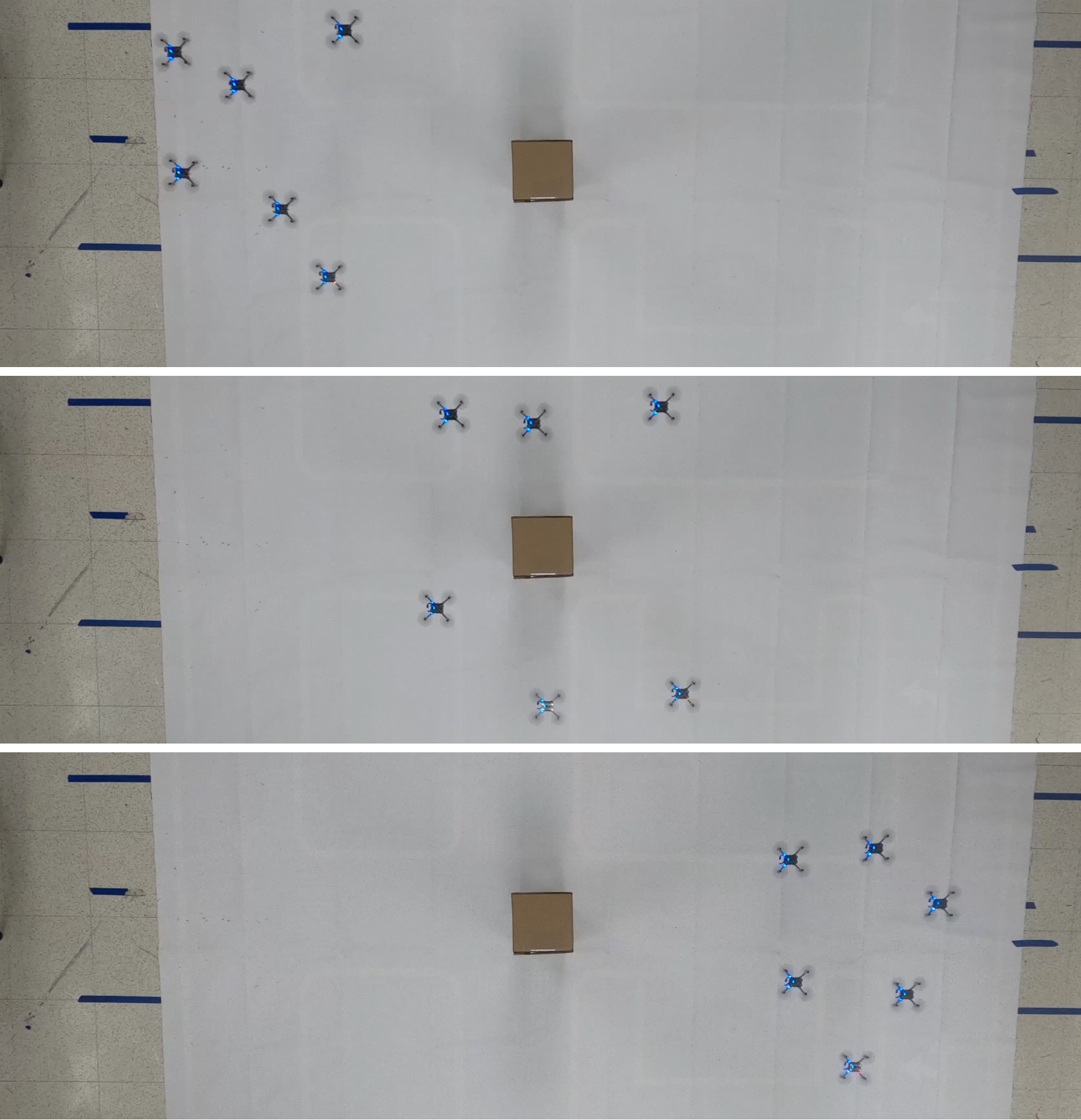}
    \caption{The plots from top to bottom show an experiment of six drones starting from random locations, avoiding an obstacle, and achieving flocking. More experiments can be found in the video attachment.}
    \label{fig:flock_exp_result}
\end{figure}

%we use crazyflie drones~\cite{crazyswarm} to implement STGNN, as show in figure~\ref{}. The positions of the drones are obtained by the lighthouse positioning system. The figure shows that it can successfully realize the flocking behaviors of the drones and avoid collision with the obstacle. The video is attached. (The video can be uploaded within 5 days after the paper submission). 
We further demonstrate the effectiveness of STGNN by implementing it to achieve flocking behaviors of a group of Bitcraze Crazyflie 2.1 drones~\cite{crazyflie}. The drones are controlled using the Crazyswarm platform~\cite{crazyswarm}, which is based on The Robot Operating System (ROS)~\cite{ros} and allows Crazyflie drones to fly as a swarm in a tight and synchronized formation. The positions of the drones are obtained using the LightHouse positioning system. The system utilizes the SteamVR base stations together with the positioning deck on the drone to estimate the position of the drones. As shown in Figure~\ref{fig:flock_exp_result}, the six drones start with random locations (top), navigate through and avoid obstacles (middle), and form a flock on the other side of the obstacle (right). 
Please refer to the online video for a more detailed real robot experiment with 4, 5, and 6 drones (see footnote 1).
%%%%%%%%%%%%%%%%%%%%%%%%%%%%%%%%%%%%%%%%%%%%%%%%%%%%%%%%%%  
\section{Conclusion and Future Work}
We demonstrate the effectiveness of STGNN as a decentralized solution for flocking with the leader following and obstacle avoidance tasks. STGNN overcomes the limitations of relying solely on local information by integrating prediction capabilities into the model, enabling it to capture and respond to global swarm dynamics.
Our STGNN-based learning model consistently outperforms the existing decentralized algorithm introduced by Olfati-Saber. Moreover, the performance of the STGNN model improves as $L$ increases. Furthermore, STGNN outperforms spatial-only models, demonstrating its ability to utilize both spatial and temporal information for enhanced flocking control.
% However, it is worth noting that in scenarios where the leader is overwhelmed by the entire swarm and the flocking behavior becomes divergent, the performance of our ST-GNN approach still falls short when compared to the expert policy.
The design is flexible in terms of $L$-hop spatial and temporal expansion, which enables seamless adaptation to various swarm sizes and history lengths. In the future, we plan to investigate the capabilities of STGNN~\cite{wu2019graph,xu2020spatial,hadou2023space} in a broader range of multi-agent tasks, including target tracking~\cite{zhou2022graph,Zhou2018ResilientAT}, path planning~\cite{li2020graph,Li2020MessageAwareGA} and coverage and exploration~\cite{tolstaya2021multi,Sharma2022D2CoPlanAD}. 
% Additionally, we intend to design learning frameworks based on different variants of STGNN that leverage spatial and temporal features 

% \clearpage
\bibliographystyle{IEEEtran}
\bibliography{refs}

\end{document}